\documentclass{article}
\usepackage{spconf,amsmath,graphicx,hyperref}
\usepackage{enumitem}
\usepackage{amsfonts}
\usepackage{float}

\title{VCE: A Zero-cost Hallucination Mitigation Method of LVLMs \\ via Visual Contrastive Editing}
\name{\begin{tabular}{@{}c@{}}
    Yanbin Huang$^{1,2\dagger}$, Yisen Li$^1$, Guiyao Tie$^1$, Xiaoye Qu$^1$ \\
    Pan Zhou$^{1\ast}$, Hongfei Wang$^{1\ast}$, Zhaofan Zou$^2$, Hao Sun$^{2\ast\ast}$, Xuelong Li$^{2\ast\ast}$
    \vspace{-3mm}
    \end{tabular}}

\address{$^1$Huazhong University of Science and Technology \\
$^2$Institute of Artificial Intelligence (TeleAI), China Telecom\\
\{m202377063, m202372098, tgy, xiaoye, panzhou, hongfei\}@hust.edu.cn \\
\{zouzhf41, sunh10\}@chinatelecom.cn, xuelong\_li@ieee.org}

\begin{document}
%
\maketitle
\renewcommand{\thefootnote}{\fnsymbol{footnote}}
\footnotetext[1]{Corresponding author.}
\footnotetext[2]{This work was completed during internship at China Telecom.}
\footnotetext[7]{The authors are with the Institute of Artificial Intelligence (TeleAI), China Telecom, P. R. China.}
\renewcommand{\thefootnote}{\arabic{footnote}}
\begin{abstract}
    Large vision–language models (LVLMs) frequently suffer from Object Hallucination (OH), wherein they generate descriptions containing objects that are not actually present in the input image.
    This phenomenon is particularly problematic in real-world applications such as medical imaging and autonomous driving, where accuracy is critical.
    Recent studies suggest that the hallucination problem may stem from language priors—biases learned during pretraining that cause LVLMs to generate words based on their statistical co-occurrence.
    To mitigate this problem, we propose Visual Contrastive Editing (VCE), a novel post-hoc method that identifies and suppresses hallucinatory tendencies by analyzing the model's response to contrastive visual perturbations.
    Using Singular Value Decomposition (SVD), we decompose the model's activation patterns to isolate hallucination subspaces and apply targeted parameter edits to attenuate its influence.
    Unlike existing approaches that require fine-tuning or labeled data, VCE operates as a label-free intervention, making it both scalable and practical for deployment in resource-constrained settings.
    Experimental results demonstrate that VCE effectively reduces object hallucination across multiple benchmarks while maintaining the model's original computational efficiency.
\end{abstract}


\section{Introduction}
\label{sec:intro}
The ability of large vision-language models (LVLMs) to interpret and contextualize visual data has unlocked transformative applications, from interactive AI assistants to real-time scene understanding systems \cite{an2026ai,luo2024transfer}.
However, their reliability is undermined by object hallucination—a tendency to generate descriptions that erroneously include objects unsupported by visual evidence \cite{Li2023EvaluatingOH}.
For instance, an LVLM might describe a person holding an umbrella in an image containing only a tree branch, a flaw with dire consequences in domains like healthcare, where misinterpreting a scan could lead to incorrect diagnoses \cite{Clusmann2025}, or in autonomous systems where hallucinated pedestrians might trigger unsafe actions \cite{8428701}.
While efforts to reduce hallucination have intensified, most solutions impose trade-offs between accuracy, computational cost, and generality, leaving a critical gap in deployable, efficient mitigation strategies \cite{math13050856}.

\begin{figure}[t]
    \centering
    \includegraphics[width=\linewidth, keepaspectratio]{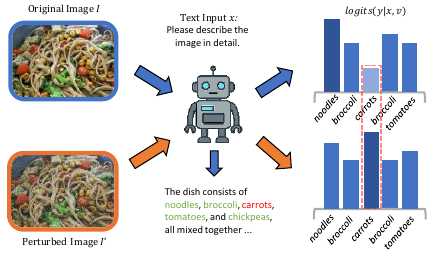}
    \vspace{-6mm}
    \caption{Perturbed images amplify language priors.}
    \vspace{-6mm}
    \label{fig:small_img}
\end{figure}
\begin{figure*}[t]
    \centering
    \includegraphics[width=\textwidth,keepaspectratio]{illustration2.pdf}
    \vspace{-6mm}
    \caption{VCE overview: Given input image $I$ and contrastively perturbed $I'$, LVLM compares logits and hidden states for token-wise confidence shifts. Aggregate and decompose via SVD to get low-rank hallucination subspace. Project model weights offline to its null space for an edited LVLM suppressing object hallucination with no extra inference cost.}
    \label{fig:vce_overview}
    \vspace{-4mm}
\end{figure*}

Existing approaches often treat hallucination as a training data imbalance or a byproduct of language-model dominance in LVLMs.
Methods like OPERA \cite{huang2024opera} rely on manually designed attention patterns to capture hallucination signals, which limits their adaptability to other scenarios.
Existing methods like TruthPrInt \cite{duan2025truthprint} and Woodpecker \cite{yin2024woodpecker} require labeled data or multiple candidate generations, limiting their use in resource-constrained scenarios.
Recent approaches such as active retrieval augmentation \cite{qu2025alleviating} and multi-view reasoning \cite{qu2025look} have shown promise in mitigating hallucinations, while multilingual hallucination mitigation remains an important challenge \cite{qu2024mitigating}.
These methods introduce validation modules or penalize implausible tokens during inference, attempting to 'patch' symptoms rather than address root causes.
Crucially, they overlook the intrinsic mechanistic origins of hallucination: deep-seated linguistic biases within pretrained LVLMs that amplify spurious correlations between objects (e.g., "plate" and "fork") regardless of visual context (Figure \ref{fig:small_img}) \cite{zhou2023analyzing}.

In this work, we propose Visual Contrastive Editing (VCE), a framework that diagnoses hallucination by synthesizing contrastive image pairs—perturbations that alter localized visual features while preserving global semantics—to isolate a hallucination subspace where the model's activations diverge when hallucinated objects are predicted.
By decomposing these activation shifts via SVD, we suppress the model's hallucination-prone components through closed-form parameter editing.
Our contributions are threefold:

\begin{itemize}[leftmargin=0.5cm, noitemsep]
    \vspace{-2mm}
    \setlength{\itemsep}{0pt}
    \item A Label-free Framework: We demonstrate that controlled visual perturbations expose the signatures of hallucination, bypassing the need for explicit hallucination labels.
    \item Subspace Model Editing: We develop a lightweight, SVD-based method to edit model parameters, reducing hallucination without post-training.
    \item Practical Efficacy: Evaluations across benchmarks show VCE effectively reduces object hallucination while preserving the model's inference speed, outperforming other methods.
\end{itemize}
\vspace{-0.4cm}
\section{METHODOLOGY}
\label{sec:format}
\subsection{Problem Definition}
Mathematically, an LVLM can be defined as a conditional probability distribution:

\begin{equation}
    p(\mathbf{y} \mid \mathbf{x}, \mathbf{v}; \theta) = \prod_{t=1}^{T} p_{\theta} \big(y_t \mid \mathbf{y}_{<t}, \mathbf{x}, \mathbf{v}\big),
\end{equation}
where \(\mathbf{y} = \{y_1, y_2, \dots, y_T\}\) is the sequence of tokens generated by the model, \(\mathbf{x}\) is the textual input, \(\mathbf{v}\) is the visual input,
\(\theta\) represents the model parameters,
\(p_{\theta} (y_t | \mathbf{y}_{<t}, \mathbf{x}, \mathbf{v})\) is the probability of generating token \(y_t\) at step \(t\), conditioned on previous tokens \(\mathbf{y}_{<t}\), the input text \(\mathbf{x}\), and the visual input \(\mathbf{v}\).

Object hallucination in LVLMs refers to the generation of objects that do not appear in the image.
Let the set of objects in the input image be denoted as \( \mathcal{O}_I = \{o_1, o_2, \dots, o_N\} \), where \( N \) is the number of objects in the image.
Let the set of predicted objects in the output text be denoted as \( \mathcal{O}_P = \{p_1, p_2, \dots, p_M\} \), where \( M \) is the number of objects predicted by the model.
An object hallucination occurs when a predicted object \( p_i \in \mathcal{O}_P \) is not part of the set of objects in the image \( \mathcal{O}_I \).
Formally, hallucination is defined as:
\begin{equation}
    \text{Hallucination} = \{p_i \mid p_i \notin \mathcal{O}_I, i=1,...,M \},
\end{equation}
where \( p_i \) is a hallucinated object.

\begin{table*}[t]
    \centering
    \caption{Results on CHAIR benchmark. Lower is better for \(\mathrm{CHAIR_S}\)/\(\mathrm{CHAIR_I}\); higher is better for BLEU. Best results in \textbf{bold}.}
    \label{tab:main-chair}
    \setlength{\tabcolsep}{4pt}
    \begin{tabular}{l|ccc|ccc|ccc}
        \hline
                                                 & \multicolumn{3}{c|}{\textbf{LLaVA-1.5}} & \multicolumn{3}{c|}{\textbf{MiniGPT-4}} & \multicolumn{3}{c}{\textbf{mPLUG-Owl2}}                                              \\
        \textbf{Method}                          & \(\mathrm{CHAIR_S}\)                    & \(\mathrm{CHAIR_I}\)                    & BLEU
                                                 & \(\mathrm{CHAIR_S}\)                    & \(\mathrm{CHAIR_I}\)                    & BLEU
                                                 & \(\mathrm{CHAIR_S}\)                    & \(\mathrm{CHAIR_I}\)                    & BLEU                                                                                 \\
        \hline
        Greedy Search
                                                 & 19.3                                    & 6.92                                    & 16.12                                   & 30.4 & 11.65 & 15.18 & 20.1 & 7.48 & 15.67 \\
        DoLa~\cite{DBLP:conf/iclr/ChuangXLKGH24} & 20.0                                    & 6.83                                    & 15.45                                   & 32.2 & 12.33 & 14.38 & 22.6 & 8.28 & 15.35 \\
        OPERA~\cite{DBLP:conf/cvpr/HuangDZ0H0L0Y24}
                                                 & 17.7                                    & 5.95                                    & 16.18                                   & 29.4 & 12.14 & 14.67 & 19.8 & 7.35 & 15.58 \\
        VCD~\cite{leng2024mitigating}            & 20.1                                    & 7.15                                    & 14.68                                   & 28.7 & 12.81 & 14.25 & 22.5 & 8.85 & 15.27 \\
        Woodpecker~\cite{yin2024woodpecker}
                                                 & 24.2                                    & 7.33                                    & 17.18                                   & 28.6 & 10.35 & 15.47 & 26.1 & 8.25 & 16.58 \\
        \hline
\textbf{VCE}
                                                 & \textbf{15.4}                           & \textbf{5.18}                           & \textbf{17.25}
                                                 & \textbf{21.7}                           & \textbf{8.76}                           & \textbf{15.52}
                                                 & \textbf{15.3}                           & \textbf{5.84}                           & \textbf{16.65}                                                                       \\
        \hline
    \end{tabular}
    \vspace{-5mm}
\end{table*}

\subsection{Constructing Contrastive Pairs}
To mitigate hallucinations without requiring manual annotation or additional data, we automatically generate \(M\) positive and negative training pairs using a contrastive approach.
Let the original dataset and perturbed dataset be defined as
\begin{equation}
    \begin{aligned}
        T^+ = \{(x_1, I_1), \dots, (x_n, I_M)\}; \\
        T^- = \{(x_1, I'_1), \dots, (x_n, I'_M)\},
    \end{aligned}
\end{equation}
where \(x_i\) is the fixed prompt, such as "describe the image in detail" and \(I_i\) is the corresponding image.

To introduce visual uncertainty, we generate a perturbed version of each image \(I_i\) using the forward diffusion process \cite{ho2020denoising}:
\begin{equation}
    q(I_t \mid I_{t-1}) = \mathcal{N}\big(I_t; \sqrt{1 - \beta_t} I_{t-1}, \beta_t \mathbf{I}\big),
\end{equation}
\begin{equation}
    q(I_{1:T} \mid I_0) = \prod_{t=1}^{T} q(I_t \mid I_{t-1}),
\end{equation}
where \(I_0 = I_i\) is the original image, \(\mathbf{I}\) is the identity matrix, \(\beta_t\) is the noise schedule at step \(t\), and \(T\) is the total number of diffusion steps.
The final perturbed image is denoted as \(I'_i = I_T\).

For each contrastive pair \((x_i, I_i), (x_i, I'_i)\),  the hidden states of each model layer \(l\) of LVLM, which \(l \in L_0,..,L\) , are generated correspondingly:
\begin{equation}
    H^{+}_{i,l} = \text{LVLM}_l(x_i, I_i), \quad
    H^{-}_{i,l} = \text{LVLM}_l(x_i, I'_i),
\end{equation}
where \(H^{+}_{i,l}, H^{-}_{i,l} =\{h_0^+,...,h_N^+\}^T, \{h_0^-,...,h_N^-\}^T \in \mathbb{R}^{N*D}\), \(N\) is the sequence length of input prompt, \(D\) is the embedding dimension, and \(h_i^+\) and \(h_i^-\) are the hidden states of the token \(t_i\) in \(T^+\) and \(T^-\), respectively.

For measuring the magnitude of language prior of each token, we compute the logit shift:
\begin{equation}
    \Delta_t = \bigl| \ell(t_i | I',x) - \ell(t_i | I,x) \bigr|,
\end{equation}
where \(\ell(t_i | I')\) is the logit of the token \(t_i\) in the perturbed image \(I'\) and \(\ell(t_i | I)\) is the logit in the original image \(I\).

We then compute a weight \(w_i\) for each token based on a robust z-scoring method.
Let $m=\mathrm{median}(\{\Delta_t\})$ and $\mathrm{MAD}=\mathrm{median}(\{|\Delta_t-m|\})$.
A robust scale is $\sigma_{\text{rob}}=1.4826\,\mathrm{MAD}$ with small $\varepsilon>0$:

\begin{equation}
    z_t \;=\; \frac{\Delta_t}{\sigma_{\text{rob}}+\varepsilon}.
\end{equation}
We map $z_t$ to a weight $w_t\in[w_{\min},1]$ via a monotone piecewise schedule:

\begin{equation}
    w_t \;=\;
    \begin{cases}
        w_{\min},                                                                  & z_t \le z_0,    \\[4pt]
        w_{\min} + (1{-}w_{\min})\!\left(\dfrac{z_t-z_0}{z_1-z_0}\right)^{\gamma}, & z_0< z_t < z_1, \\[10pt]
        1,                                                                         & z_t \ge z_1.
    \end{cases}
\end{equation}
Unless stated otherwise, we use $z_0{=}1.5$, $z_1{=}3.5$, $\gamma{=}2.0$, $w_{\min}{=}0.05$.

The final editing vector \(v\) is constructed by aggregating the weighted differences of the hidden states of tokens in \(T^+\) and \(T^-\):
\begin{align}
    v_d & = \frac{1}{N} \sum_{i=1}^{N} w_i \cdot (h_i^- - h_i^+),
\end{align}
where \(d = 0,... , M\).
Then the language prior subspace of \(l\) model layer can be modeled as \(V_l=\{v_0,...,v_M\}^T \in \mathbb{R}^{M*D}\).
This vector is derived from the hidden state differences between the positive and negative samples, where tokens with high logit differences are deemed more likely to be hallucinated and are weighted more heavily.

\subsection{Hallucination Subspace Discovery}
To identify dominant hallucination directions, we perform Singular Value Decomposition (SVD):
\begin{equation}
    \mathbf{V}_l = U_l \Sigma_l S_l^\top,
\end{equation}
where $U_l \in \mathbb{R}^{M \times M}$ contains orthogonal basis vectors, $\Sigma_l$ is a diagonal matrix, and $S_l \in \mathbb{R}^{D \times D}$ is an orthogonal matrix.
The top-$k$ singular vectors corresponding to the largest $\sigma_{li}$ form the hallucination subspace (HalluSpace) for layer $l$:
\begin{equation}
    S_l^k = [s_{l1}, s_{l2}, \dots, s_{lk}] \in \mathbb{R}^{D \times k}.
\end{equation}
We then suppress hallucination by projecting the model weights $W_l$ to the null space:
\begin{equation}
    W_l^{\text{edited}} = W_l \left(\mathbf{I} - S_l^k (S_l^k)^\top \right),
\end{equation}
where \(\mathbf{I}\) is the identity matrix.
This offline editing eliminates hallucination-prone components without inference overhead.
Overview of VCE is depicted in Figure \ref{fig:vce_overview}.

\begin{table*}[ht]
    \centering
    \caption{POPE results on object presence evaluation. Higher is better. Best results in \textbf{bold}.}
    \label{tab:opope}
    \setlength{\tabcolsep}{6pt}
    \begin{tabular}{l|ccc|ccc|ccc}
        \hline
                                                 & \multicolumn{3}{c|}{\textbf{LLaVA-1.5}} & \multicolumn{3}{c|}{\textbf{MiniGPT-4}} & \multicolumn{3}{c}{\textbf{mPLUG-Owl2}}                                                 \\
        \textbf{Method}                          & Accuracy                                & Precision                               & F1
                                                 & Accuracy                                & Precision                               & F1
                                                 & Accuracy                                & Precision                               & F1                                                                                      \\
        \hline
        Greedy Search
                                                 & 79.18                                   & 92.67                                   & 91.13                                   & 71.48 & 94.55 & 90.74 & 76.83 & 90.15 & 88.43 \\
        DoLa~\cite{DBLP:conf/iclr/ChuangXLKGH24} & 79.15                                   & 91.43                                   & 90.28                                   & 71.45 & 93.67 & 90.35 & 76.22 & 88.31 & 87.12 \\
        OPERA~\cite{DBLP:conf/cvpr/HuangDZ0H0L0Y24}
                                                 & 79.46                                   & 92.08                                   & 90.58                                   & 70.35 & 94.28 & 90.43 & 75.67 & 91.08 & 89.28 \\
        VCD~\cite{leng2024mitigating}            & 78.28                                   & 91.18                                   & 89.56                                   & 70.65 & 92.48 & 88.63 & 75.67 & 88.52 & 87.19 \\
        Woodpecker~\cite{yin2024woodpecker}
                                                 & 79.46                                   & 92.08                                   & 90.58                                   & 70.35 & 94.28 & 90.43 & 75.67 & 91.08 & 89.28 \\
        \hline
        \textbf{VCE}
                                                 & \textbf{79.67}                          & \textbf{93.23}                          & \textbf{91.56}
                                                 & \textbf{72.15}                          & \textbf{95.73}                          & \textbf{91.84}
                                                 & \textbf{77.34}                          & \textbf{92.65}                          & \textbf{90.67}                                                                          \\
        \hline
    \end{tabular}
    \vspace{-3mm}
\end{table*}

\begin{table}[t]
    \centering
    \small
    \setlength{\tabcolsep}{4pt}
    \caption{Complexity \& Efficiency Ablations on LLaVA-1.5}
    \label{tab:eff}
    \begin{tabular}{lccc}
        \hline
        Method        & Complexity        & per answer \\
        \hline
        Greedy Search & \(O(n^2)\)        & 7.2s           \\
        DoLa          & \(O(n^2+\alpha)\) & 7.8s           \\
        OPERA         & \(O((n+k)^2)\)    & 9.4s           \\
        VCD           & \(O(2n^2)\)       & 14.1s          \\
        VCE           & \(O(n^2)\)        & 7.2s           \\
        \hline
    \end{tabular}
    \vspace{-3mm}
\end{table}

\begin{table}[t]
    \centering
    \small
    \setlength{\tabcolsep}{4pt}
    \caption{Hyperparameter Ablations on LLaVA-1.5}
    \label{tab:ablate}
    \begin{tabular}{lccc}
        \hline
        Setting             & \(C_S\)       & \(C_I\)       & BLEU          \\
        \hline
        $\ell\!=\!16$--$32$ & \textbf{15.4} & \textbf{5.12} & \textbf{15.9} \\
        $\ell\!=\!24$--$32$ & 16.0          & 5.43          & 15.3          \\
        $\ell\!=\!28$--$32$ & 18.5          & 6.15          & 14.8          \\
        $\ell\!=\!31$--$32$ & 16.7          & 5.34          & 15.4          \\
        \hline
        $k\!=\!2$           & 20.0          & 7.18          & 15.1          \\
        $k\!=\!4$           & \textbf{15.4} & \textbf{5.12} & \textbf{15.9} \\
        $k\!=\!8$           & 17.6          & 5.78          & 15.0          \\
        $k\!=\!16$          & 17.4          & 6.33          & 14.6          \\
        \hline
    \end{tabular}
    \vspace{-3mm}
\end{table}

\section{EXPERIMENTS}
\label{sec:pagestyle}
\subsection{Experiment Setups}

\noindent\textbf{Datasets.} We evaluate the VCE method on popular hallucination benchmarks: POPE \cite{li2023evaluating} and  CHAIR \cite{rohrbach-etal-2018-object}.
Each dataset consists of questions, images and answers.

\noindent\textbf{LVLM Models.} To test the generalization of our method, We select three commonly used LVLMs:  LLaVA-1.5 \cite{liu2024improved}, MiniGPT-4 \cite{DBLP:conf/iclr/Zhu0SLE24} and mPLUG-Owl2 \cite{ye2024mplug}.

\noindent\textbf{Evaluation Metrics.} We evaluate the object hallucination rate using \(\mathrm{CHAIR_s}\) and \(\mathrm{CHAIR_I}\) metrics \cite{rohrbach-etal-2018-object}.
And we also utilize the accuracy, precision, F-score and BLEU to evaluate the generation quality.

\noindent\textbf{Settings.} All our experiments are conducted on an NVIDIA RTX 5090 card.
Each experiment is repeated three times with different random seeds.
\vspace{-3mm}
\subsection{Main Performance}
\label{sec:main_results}

We evaluate the performance of our method alongside a suite of object hallucination mitigation strategies in LVLMs.
Our method (VCE) significantly reduces object hallucination across all evaluated LVLMs.
On CHAIR dataset (Table~\ref{tab:main-chair}), VCE achieves the best \(\mathrm{CHAIR_S}\), \(\mathrm{CHAIR_I}\) and BLEU (i.e., best hallucination suppression and generation quality) for all three models (LLaVA-1.5, MiniGPT-4, mPLUG-Owl2), outperforming all baselines.
For POPE object grounding (Table~\ref{tab:opope}), VCE delivers highest accuracy, precision and F1 across all models, confirming consistent improvements in object presence detection.
These results demonstrate VCE's effectiveness in reducing hallucinations while maintaining caption quality.
\vspace{-3mm}
\subsection{Ablation Study}
\textbf{Layers to edit.} VCE has two key hyper-parameters: LVLM layers $\{\ell\}$ to edit and the subspace rank $k$.
As shown in Table \ref{tab:ablate}, editing a mid-to-deep block delivers the most favorable CHAIR trade-off.
Restricting edits to only the last few layers can under-correct, whereas too shallow a range may over-regularize.

\noindent\textbf{Rank $k$ of SVD.} In Table \ref{tab:ablate}, we observe a clear sweet spot at small ranks, consistent with low-rank structure in truthful vs. hallucinated feature differences.
For LLaVA-1.5, $k\!=\!4$ performs best.
This supports the intuition that a compact subspace captures the dominant hallucination directions while avoiding collateral loss of truthful semantics.

\noindent\textbf{Efficiency \& time complexity.} Owing to the auto-regressive nature of LVLMs, the time complexity of generating a sequence of length $n$ is $O(n^2)$.
Methods such as VCD \cite{leng2024mitigating}, which perform two inference passes before generation, exhibit a complexity of $O(2n^2)$.
In contrast, the proposed VCE method introduces no additional overhead during inference, making it more computationally efficient.
Inference overhead of other methods are listed in Table \ref{tab:eff}.

\vspace{-3mm}
\section{Conclusion}
\vspace{-3mm}
\label{sec:conclusion}
In this work, we presented Visual Contrastive Editing (VCE), a zero-cost, post-hoc framework for mitigating object hallucination in large vision–language models.
By leveraging contrastive visual perturbations to expose confidence shifts and employing singular value decomposition to isolate a low-rank hallucination subspace, VCE performs targeted parameter editing that suppresses spurious activations without additional training data, fine-tuning, or inference overhead.
Owing to its model-agnostic design, VCE offers a scalable, practical solution for improving the reliability of multimodal large language models and can be readily extended to other modalities and application domains.
\label{sec:acknowledge}

\let\oldthebibliography\thebibliography
\renewcommand{\thebibliography}[1]{%
  \oldthebibliography{#1}%
  \small  
}
\bibliographystyle{IEEEbib}
\bibliography{strings,refs}

\end{document}